\documentclass[sigconf]{acmart}
\usepackage{booktabs} % For formal tables
\newenvironment{myitemize}[1][]{%%%%%
	\begin{list}{$\bullet$}
		{
			\setlength{\leftmargin}{5mm}
			\setlength{\parsep}{1mm}
			\setlength{\topsep}{0mm}
			\setlength{\itemsep}{0mm}
			\setlength{\labelsep}{1.5mm}
			\setlength{\itemindent}{0mm}
			\setlength{\listparindent}{5mm}
	}}
	{\end{list}}%%%%%

\usepackage{subfigure}

\usepackage{balance}
\def\BibTeX{{\rm B\kern-.05em{\sc i\kern-.025em b}\kern-.08emT\kern-.1667em\lower.7ex\hbox{E}\kern-.125emX}}

\copyrightyear{2020}
\acmYear{2020}
\setcopyright{acmcopyright}\acmConference[MM '20]{Proceedings of the 28th ACM
International Conference on Multimedia}{October 12--16, 2020}{Seattle, WA, USA}
\acmBooktitle{Proceedings of the 28th ACM International Conference on Multimedia
(MM '20), October 12--16, 2020, Seattle, WA, USA}
\acmPrice{15.00}
\acmDOI{10.1145/3394171.3414031}
\acmISBN{978-1-4503-7988-5/20/10}

\begin{document}

\fancyhead{}
  % do not delete this code.

% The "title" command has an optional parameter, allowing the author to define a "short title" to be used in page headers.
\title{ISIA Food-500: A Dataset for Large-Scale Food Recognition via Stacked Global-Local Attention Network}

\author{Weiqing Min$^{1,2}$, Linhu Liu$^{1,2}$, Zhiling Wang$^{1,2}$, Zhengdong Luo$^{1,2}$, Xiaoming Wei$^{3}$, Xiaolin Wei$^{3}$, Shuqiang Jiang$^{1,2}$}

\affiliation{%
  \institution{$^1$ Key Lab of Intelligent Information Processing, Institute of Computing Technology, CAS, Beijing, 100190, China}
  \institution{$^2$ University of Chinese Academy of Sciences, Beijing, 100049, China}
  \institution{$^3$ Meituan-Dianping Group}
}
\email{{minweiqing,sqjiang,luozhengdong}@ict.ac.cn; {linhu.liu,zhiling.wang}@vipl.ict.ac.cn;{weixiaoming,weixiaolin02}@meituan.com}
%\email{{minweiqing,sqjiang,luozhengdong}@ict.ac.cn; {linhu.liu,zhiling.wang}@vipl.ict.ac.cn; }
%\email{{weixiaoming,weixiaolin02}@meituan.com}

%\author{Weiqing Min}
%\affiliation{%
%  \institution{Key Lab of Intelligent Information Processing, Institute of Computing Technology, CAS}
%  \institution{University of Chinese Academy of Sciences}
%  \city{Beijing}
%\email{minweiqing@ict.ac.cn}
%
%\author{Linhu Liu}
%\affiliation{%
%  \institution{Key Lab of Intelligent Information Processing, Institute of Computing Technology, CAS}
%  \institution{University of Chinese Academy of Sciences}
%  \city{Beijing}
%\email{linhu.liu@vipl.ict.ac.cn}
%
%\author{Zhiling Wang}
%\affiliation{%
%  \institution{Key Lab of Intelligent Information Processing, Institute of Computing Technology, CAS}
%  \institution{University of Chinese Academy of Sciences}
%  \city{Beijing}
%\email{zhiling.wang@vipl.ict.ac.cn}
%
%\author{Zhengdong Luo}
%\affiliation{%
%  \institution{Key Lab of Intelligent Information Processing, Institute of Computing Technology, CAS}
%  \institution{University of Chinese Academy of Sciences}
%  \city{Beijing}
%\email{luozhengdong@ict.ac.cn}
%
%
%\author{Shuqiang Jiang}
%\affiliation{%
%  \institution{Key Lab of Intelligent Information Processing, Institute of Computing Technology, CAS}
%  \institution{University of Chinese Academy of Sciences}
%  \city{Beijing}
%\email{sqjiang@ict.ac.cn}

%
% The abstract is a short summary of the work to be presented in the article.
\begin{abstract}
Food recognition has received more and more attention in the multimedia community for its various real-world applications, such as diet management and self-service restaurants. A large-scale ontology of food images is urgently needed for developing advanced large-scale food recognition algorithms, as well as for providing the  benchmark dataset for such algorithms.  To encourage further progress in food recognition, we introduce the dataset ISIA Food-500 with  500 categories from the list in the Wikipedia and 399,726 images, a more comprehensive food dataset that surpasses  existing popular benchmark datasets by category coverage and data volume. Furthermore, we propose a stacked global-local attention network, which  consists of  two sub-networks  for food recognition. One sub-network  first utilizes hybrid spatial-channel attention to extract more discriminative features, and then aggregates these multi-scale discriminative features from multiple layers into global-level  representation  (e.g., texture and shape information about food). The other one  generates attentional regions (e.g., ingredient relevant regions) from different regions  via cascaded spatial transformers, and further aggregates these multi-scale regional features from different layers  into local-level representation. These two types of features are finally fused as comprehensive representation for food recognition. Extensive experiments on ISIA Food-500 and other two popular benchmark datasets  demonstrate the  effectiveness of our proposed method, and thus can be considered as one strong baseline. The dataset, code and models can be found at {\color{red}{\url{http://123.57.42.89/FoodComputing-Dataset/ISIA-Food500.html}}}.
\end{abstract}



%
% Keywords. The author(s) should pick words that accurately describe the work being
% presented. Separate the keywords with commas.

%
% This command processes the author and affiliation and title information and builds
% the first part of the formatted document.
\maketitle

\section{Introduction}

Food computing~\cite{Min2019A} is emerging as a new field to ameliorate the issues from many food-relevant fields, such as nutrition, agriculture and medicine. As one significant task in food computing, food recognition has received more attention in multimedia and beyond\\~\cite{Kagaya-FDR-MM2014,Meyers-Im2Calories-ICCV2015,Deng-MDR-MM2019,Min-IGCMAN-ACMMM2019} for its various applications, such as visual food diary~\cite{Meyers-Im2Calories-ICCV2015}, health-aware recommendation~\cite{Nag-HML-ICMR2017} and self-service restaurants~\cite{Aguilar2018Grab}.

\begin{figure}[h]
	\centering
	\includegraphics[width=0.45\textwidth]{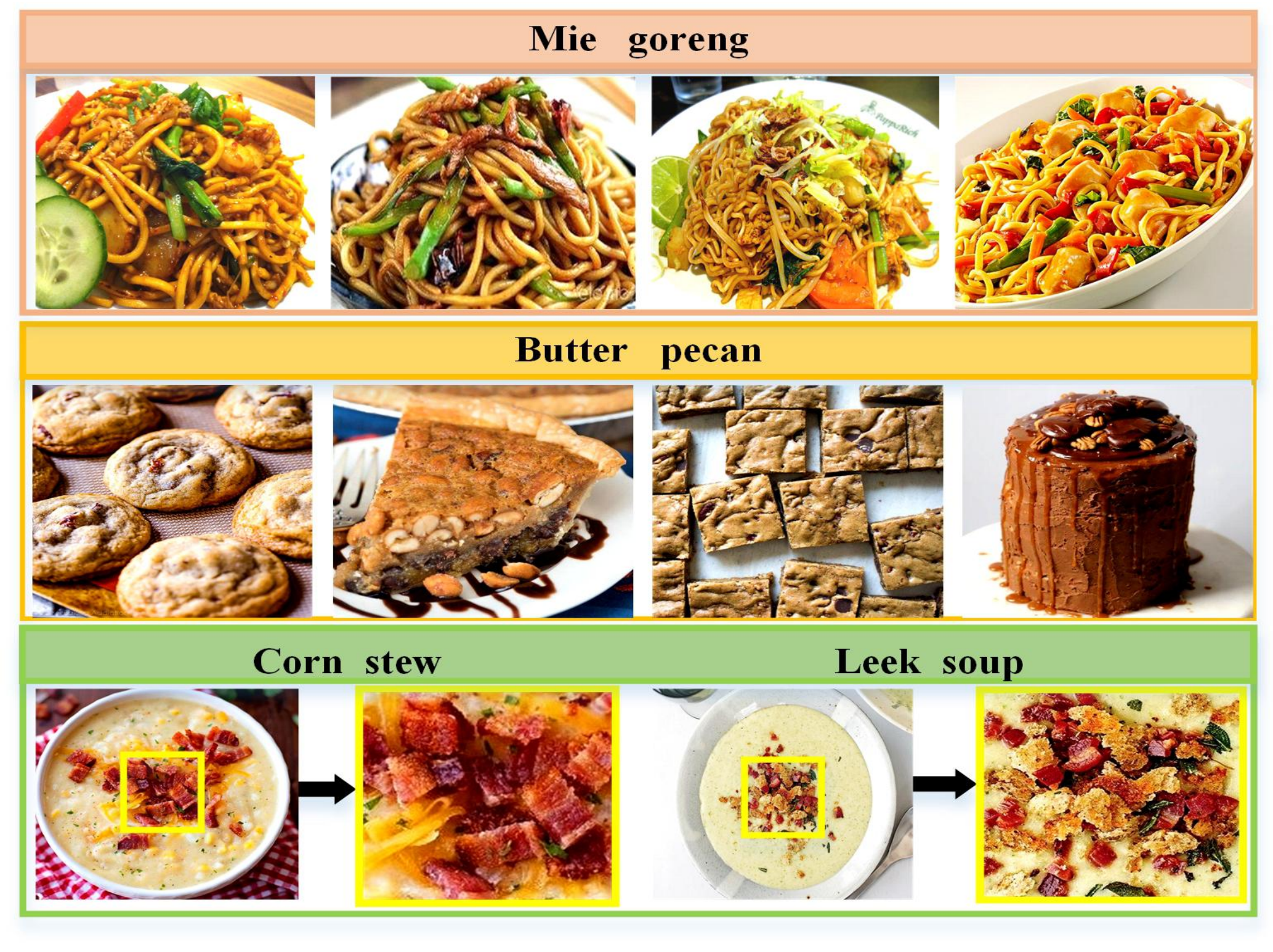}
	\caption{Some samples from ISIA Food-500}
	\label{samples_intro}
\end{figure}

Despite its great potential applications, recognizing food from images is still a challenging task, and  the challenge derives from three-fold:
\begin{myitemize}
	\item \textbf{There is a lack of large-scale food dataset for food recognition.} Existing works  mainly focus on utilizing  smaller  datas-ets for food recognition, such as ETH Food-101~\cite{Bossard-Food101-ECCV2014} and Vireo Food-172~\cite{Chen-DIRCRR-MM2016}. For example, Bossard \textit{et al.}~\cite{Bossard-Food101-ECCV2014} released one food dataset ETH Food-101  from western cuisines  with 101 food categories and 101,000 images. Chen \textit{et al.}~\cite{Chen-DIRCRR-MM2016} introduced the Vireo Food-172  dataset from 172 Chinese food categories.  These data-sets is lack of diversity and coverage in food categories and do not include a wide range of food images. Therefore, they are probably not sufficient to construct more complicated deep learning models for food recognition.
	\item \textbf{There are  larger intra-class variations in the global appearance, shape and other configurations for food images.} As shown in Fig.~\ref{samples_intro}, there are different shapes for the butter pecan and different textures appear in the mie goreng dish.  Although numerous methods have been developed for addressing the problem of food recognition,  most of these methods mainly focus on extracting features with certain type or some types while ignoring other aspects. For example, works on ~\cite{Bettadapura-LCSAFR-WACV2015} mainly extracted color features  while  Niki \textit{et al.} ~\cite{Martinel-WSR-WACV2018} designed a network to  capture  certain vertical structure for food recognition.
	\item \textbf{There are subtle discriminative details from food images, which are harder to  capture in many cases.}  Food recognition belongs to fine-grained recognition. Therefore,  discriminative details are too subtle to be well-represented by  existing CNNs in many cases. As shown in Fig.~\ref{samples_intro}, global features are not  discriminative enough to distinguish between corn stew and leek soup. Although  local regional features are probably more useful, we should carefully design one network to capture and represent such subtle difference. In order to improve the recognition performance,  additional context information, such as location and ingredients~\cite{Bettadapura-LCSAFR-WACV2015,XuRuihan-GMDR-TMM2015,Zhou_FGIC_CVPR2016,Min-IGCMAN-ACMMM2019} is utilized. However, when  these information is unavailable, these methods probably do not work.
\end{myitemize}

In this work, we address data limitations by introducing a new large-scale  dataset ISIA Food-500 with 399,726 images and 500 categories. In contrast with existing popular benchmark datasets, it is a more comprehensive food dataset with larger category coverage, larger data volume and  higher diversity. To solve another two challenges, we propose a Stacked Global-Local Attention Network (SGLANet) to jointly learn complementary global and local visual features for food recognition. This is achieved by two sub-networks, namely Global Feature Learning Subnetwork~(GloFLS) and Local-Feature Learning Subnetwork~(LocFLS). GloFLS  first utilizes hybrid spatial-channel attention to obtain more discriminative  features for each layer, and then  aggregates these  features  from different  layers  with both coarse and fine-grained levels, such as shape and texture cues about food into global-level features. LocFLS adopts cascaded Spatial Transformers~(STs) to localize different  attentional regions (e.g., ingredient-relevant regions), and   aggregates fused regional features from different layers into  local-level representation. In addition, SGLANet  is trained  with different types of losses in an end-to-end fashion to maximize their complementary effect in terms of discriminative power.

The contributions of our paper can be summarized as follows:

\begin{myitemize}
	\item We introduce a new large-scale and highly diverse food image dataset with  500 categories and about 400,000 images, which will be made publicly available to further the development of scalable  food recognition.
	\item We propose  a  stacked global-local attention network  architecture to jointly learn food-oriented global and local features via combining hybrid spatial-channel attention and multi-scale strategy for food recognition.
	\item We conduct extensive  evaluation on our proposed dataset and other two popular food benchmark datasets to verify the effectiveness of  our approach.  As one strong baseline, code and models  will also be released upon publication to support future research.
\end{myitemize}
\section{Related Work}
\begin{table}[t]	
	\caption{Summary of available datasets for food recognition.}
	\label{datasets_summary}
	\small
	\begin{center}
		\begin{tabular}{cccc}
			\hline
			Dataset&\#Images&\#Categories&\#Coverage\\
			\hline
			PFID~\cite{Chen2009PFID} &4,545& 101 &Japanese\\
			UEC Food100~\cite{Matsuda2012Multiple} &14,361 & 100 &Japanese\\
			UEC Food256~\cite{kawano2014automatic} &25,088& 256&Japanese\\
			ETHZ Food-101~\cite{Bossard-Food101-ECCV2014} &101,000& 101&Western\\
			UPMC Food-101~\cite{Wang-RRLMFD-ICME2015}& 90,840&101&Western\\
			UNIMIB2015~\cite{Ciocca2015Food}& 2,000&15&Misc.\\
			UNIMIB2016~\cite{Ciocca2016Food}& 1,027&73&Misc.\\
			ChineseFoodNet~\cite{Chen2017ChineseFoodNet}&192,000&208&Chinese\\
			Vireo Food-172~\cite{Chen-DIRCRR-MM2016}&110,241&172&Chinese\\
			KenyanFood13~\cite{Jalal-SSMPP-MADiMa2019}&8,174&13&Kenyan\\
			Sushi-50~\cite{Qiu-MDFR-BMVC2019}&3,963&50&Japanese\\
			FoodX-251~\cite{Parneet-FoodX251-CVPRW2019}&158,846&251&Misc.\\
			ISIA Food-200~\cite{Min-IGCMAN-ACMMM2019}&197,323&200&Misc.\\
			\hline
			ISIA Food-500&\textbf{399,726}&\textbf{500}&\textbf{Misc.}\\
			\hline
		\end{tabular}
	\end{center}
	
\end{table}
\textbf{Food-centric datasets} More and more food datasets  have been developed~\cite{Matsuda2012Multiple,kawano2014automatic,Bossard-Food101-ECCV2014,Chen-DIRCRR-MM2016,Parneet-FoodX251-CVPRW2019,Min-IGCMAN-ACMMM2019} in recent years. Table \ref{datasets_summary} summarizes statistics of publicly available datasets for food recognition. The first benchmark is the PFID dataset~\cite{Chen2009PFID} with only 4,545 images from 101 fast food categories. ETHZ Food-101 dataset~\cite{Bossard-Food101-ECCV2014}  and VIREO Food-172 dataset~\cite{Chen-DIRCRR-MM2016} consist of more food images. However, these datasets failed in term of more comprehensive coverage of food categories, like object-centric ImageNet~\cite{Deng-ImageNet-CVPR2009} and place-centric Places~\cite{Bolei-Places-TPAMI2018}. We hence introduce a new large scale food dataset ISIA Food-500 with 399,726 images and 500 food categories, and it aims at advancing multimedia food recognition and promoting the development of food-oriented multimedia intelligence.

There are some recipe-relevant multimodal datasets, such as Yummly28K~\cite{Min2017Being}, Yummly66K~\cite{Min-YAWYE-TMM2018} and Recipe1M~\cite{Salvador-LCME-CVPR2017}. Recipe1M is the most known dataset, which  contains about 1 million structured cooking recipes and their images  for cross-modal retrieval.  In contrast, the goal of our proposed ISIA Food-500 is for  advancing multimedia food recognition.

\textbf{Food Recognition} Recently, Min \textit{et al.}~\cite{Min2019A} gave a  survey on food computing including food recognition. In the earlier years, various hand-crafted features are utilized  for recognition~\cite{Yang-FR-CVPR2010,Bossard-Food101-ECCV2014}. For example, Lukas \textit{et al.}~\cite{Bossard-Food101-ECCV2014} utilized random forests to mine discriminative image patches as visual representation. Recent advances in deep learning  have gained significant attention due to its impressive performance. As a result, existing methods resort to deep learning for food recognition~\cite{Kagaya-FDR-MM2014,Martinel-WSR-WACV2018,Horiguchi-PCFIR-TMM2018}. There are also literatures, which  utilize additional context information, such as ingredients and location~\cite{Chen-DIRCRR-MM2016,Zhou_FGIC_CVPR2016,Min-IGCMAN-ACMMM2019} to improve the recognition performance.
For example, Zhou \textit{et al.}~\cite{Zhou_FGIC_CVPR2016} exploited rich relationships among ingredients and restaurant information through the bi-partite graph for food recognition. Different from these works, our work does not introduce additional context information, and design a two-branch network to jointly learn food-oriented global  features(e.g., texture and shape) and local features (e.g.,ingredient-relevant regional features) to enable comprehensive and discriminative  feature representation for food recognition.

In addition, our work is also very relevant to fine-grained image recognition~\cite{Shen-FGIA-arXiv2019}, which  aims to classify subordinate categories.  Food recognition belongs to fine-grained image recognition. However, compared with  other types of fine-grained objects, we should take  characteristics of food images into consideration, and design  the targeted network for food recognition.
\section{ISIA Food-500}
\begin{figure*}
	\centering
	\includegraphics[width=1.0\textwidth]{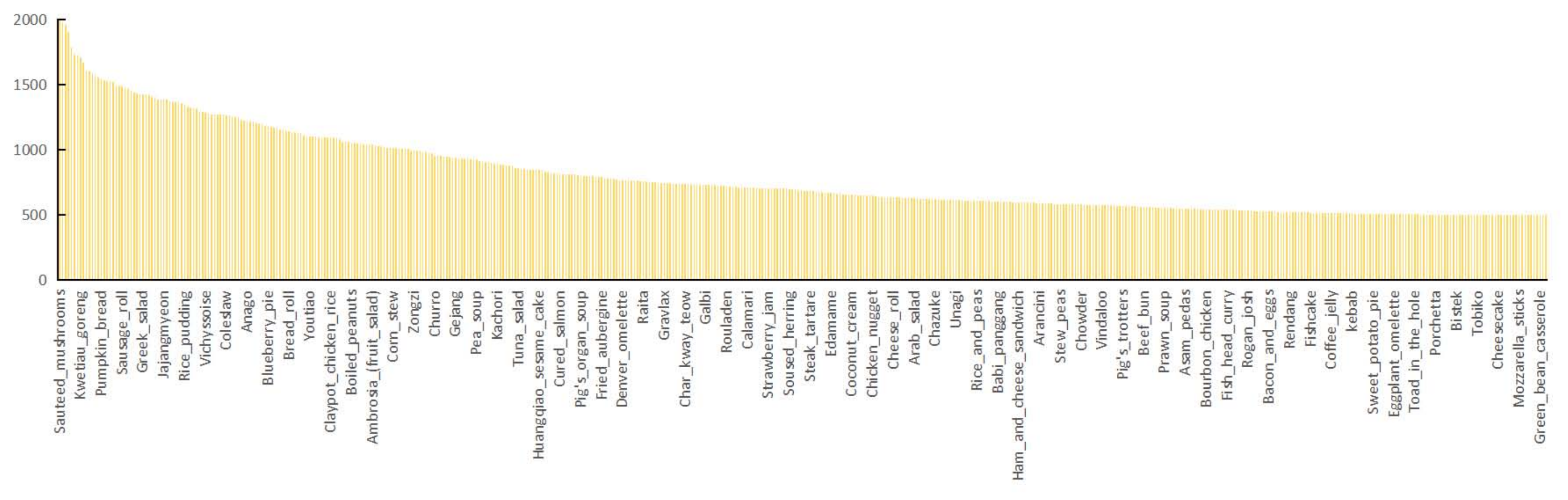}
	\caption{Sorted distribution of the number of images from sampled classes in the ISIA Food-500.}
	\label{distribution}
\end{figure*}

\subsection{Dataset Construction}
In order to obtain one high-quality food dataset with broad coverage, high diversity and density of samples, we build ISIA Food-500 from the following four steps:

\textbf{(1) Constructing the Food Category List.} In order to guarantee high-coverage of the categorical space, we resort to Wikipedia to construct the food concept system. Particularly, we  built the food list according to  ``Lists of foods by ingredient" from  Wikipedia\footnote{\url{https://en.wikipedia.org/wiki/Category:Lists_of_foods}}. The Deep-First-Search algorithm is used to  traverse links of the website to find food categories more completely. After that, we obtained the original food list with 4,943 types. We then removed redundant food types and conducted the combination for synonyms. Finally, we obtained 3,309  food categories.

\textbf{(2) Collecting Food Images.} Using a query term from the constructed food category list, we crawled candidate images from various  search engines (i.e., Google, Bing and Baidu) for broader coverage and higher diversity of food images compared with other datasets from only one data source. In order to ensure that crawled images are less noisy, we expanded search terms by adding keywords, such as ``food" and ``dish". In this case, images for each term are retrieved and  these images are then combined from different search engines. Because some images crawled from different search engines are repeated, we conducted hash based duplication detection to remove repeated ones.

\textbf{(3) Cleaning and Pre-processing Food Images.} Images are cleaned up through both automatic and manual processing. For automatic data cleaning, we removed candidate images with incomplete RGB channels, and the length or width of an image less than 100 pixels. We next trained a food/non-food binary classifier to further remove non-food images. Particularly, we combined images from the training set of both ETHZ Food-101 (western dishes) and VireoFood-172 (eastern dishes) as positive samples of the training set. We then randomly selected about 400,000 non-food images from both ImageNet and Places365 as negative samples of the training set. All the test samples of both ETHZ Food-101 and VireoFood-172 and the other 100,000 non-food images randomly selected from both ImageNet and Places365 constitute the test set. We trained a  deep network (VGG-16 in our work) on the constructed training set and the classification accuracy of the network achieved 99.48\% on the test set. The trained model is then used to filter out non-food images from downloaded images. After automatic cleaning, we then conduct manual  verification by crowd-sourcing the task to 20 Lab members.

\textbf{(4) Scaling Up the Dataset.} After image collection and annotation, there are still many food categories with few images. To further increase the number of the candidate dataset, we translated the name of these food categories into different languages, such as Chinese and French, and then crawled  images from three search engines. We also crawled more food images from other recipe/food shared websites, such as Allrecipes.com  and foodgawker.com. We finally selected 500 categories with more than 500 images per category  as our resulting  dataset.
\begin{figure}
	\centering
	\includegraphics[width=0.45\textwidth]{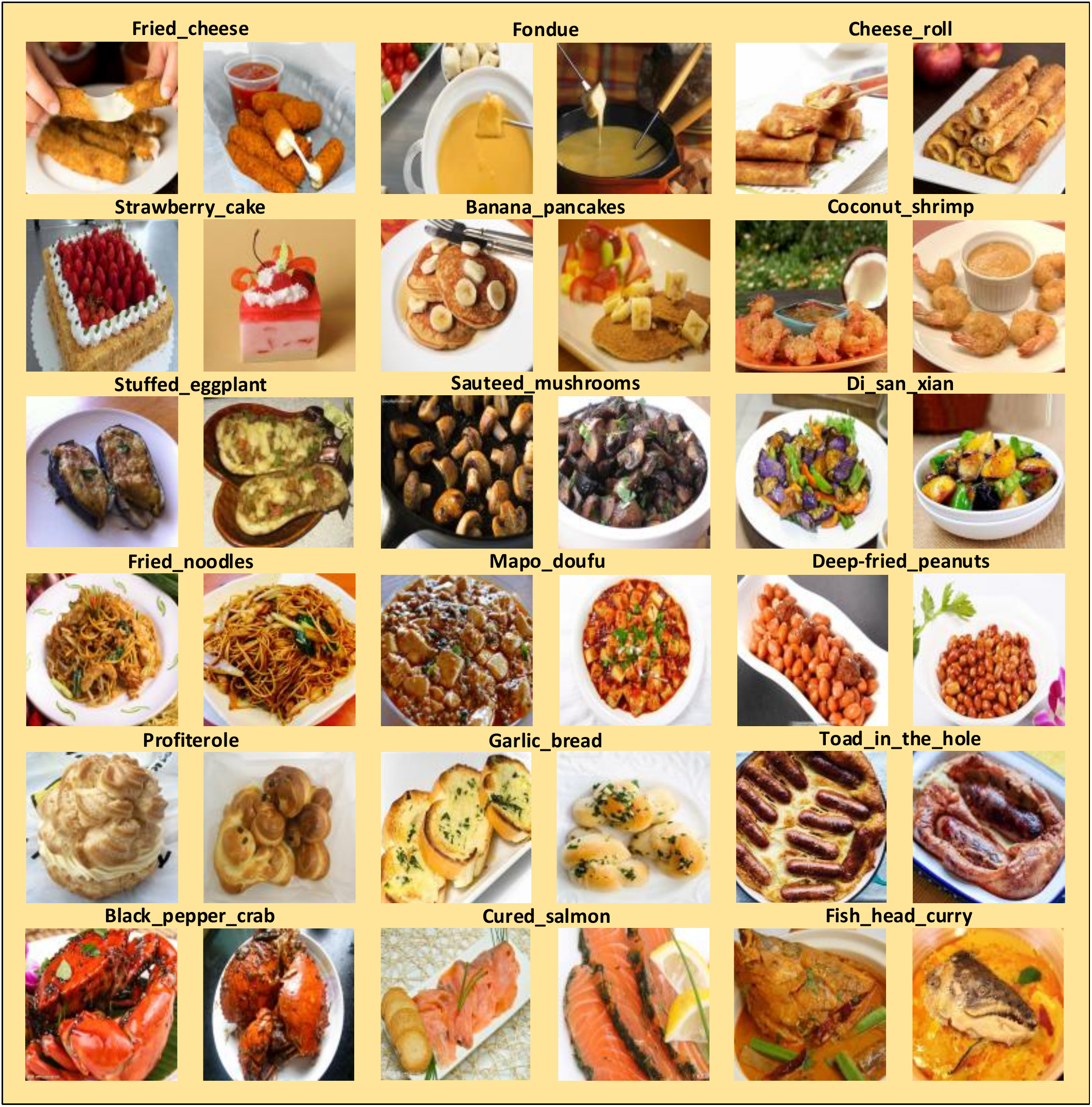}
	\caption{Image samples from the ISIA Food-500 dataset}
	\label{samples}
\end{figure}
\begin{figure*}
	\centering
	\includegraphics[width=0.9\textwidth]{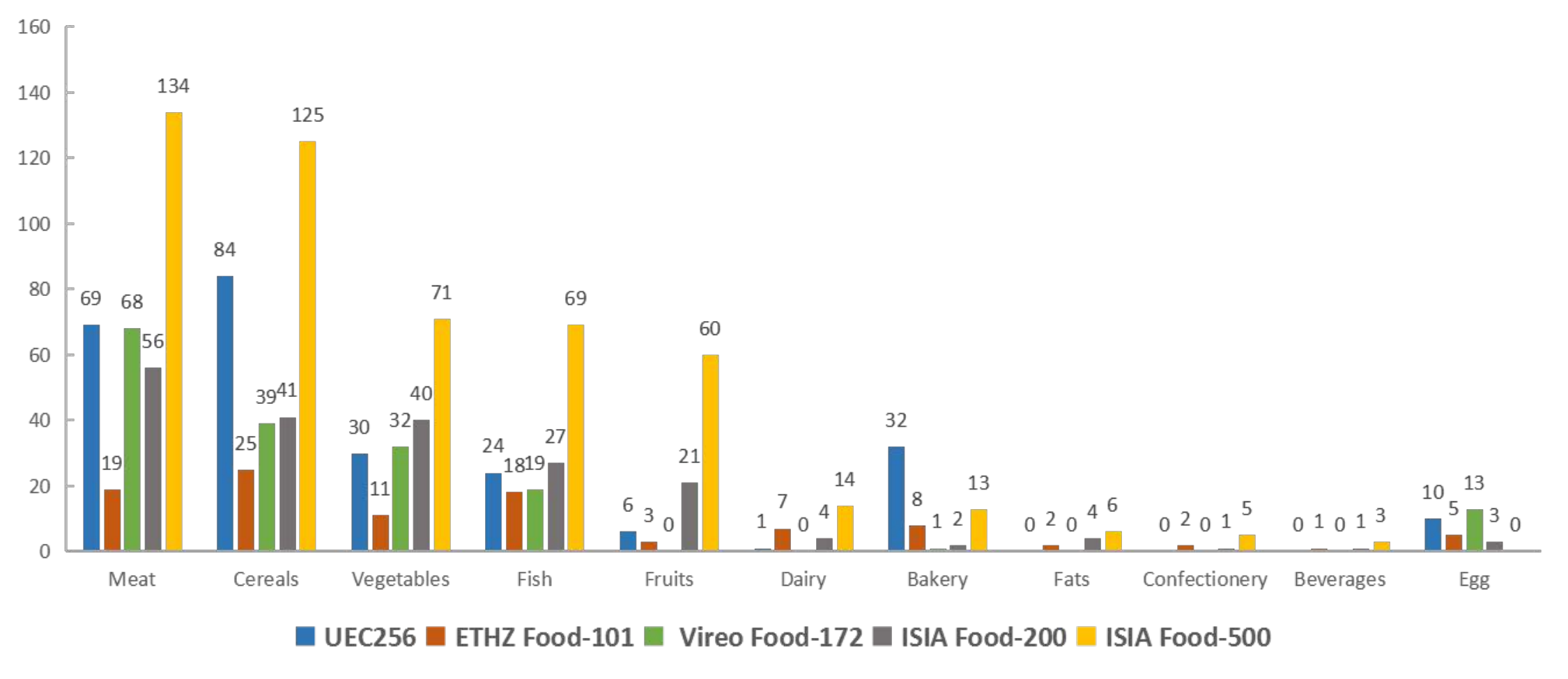}
	\caption{Comparison on distributions of categories on ISIA Food-500 and other existing typical
		ones.}
	\label{food_dataset_comparison}
\end{figure*}

\subsection{Dataset Statistics and Characteristics}
\label{data_sta}
ISIA Food-500 consists of 399,726 images with 500 categories.  The average number of images per category is about 800. Fig. \ref{distribution} shows sorted distribution  of the number of images from sampled classes while Fig. \ref{samples} shows some samples. Note that we represented the food category with more than two words by concatenating them using `-'. ISIA Food-500 is a more comprehensive food dataset that surpasses  existing popular benchmark datasets, such as ETH Food-101 and Vireo Food-172 from the following three aspects: \textbf{(1) Larger data volume}. It has 399,726 images from 500 food categories, which has created a new milestone for the task of complex food recognition. \textbf{(2) Larger category coverage}. It consists of 500 categories, which is about $3\sim5$ times that of existing datasets, such as Food-101 and Vireo Food-172. \textbf{(3) Higher diversity}.  Food categories from this dataset covers  various countries and regions including both eastern and western cuisines.  Fig. \ref{food_dataset_comparison} provided the comparisons of distributions of food categories on food types, such as ETH Food-101 (western food), Vireo Food-172 (eastern food) and ISIA Food-200 (Misc. food).  According to the GSFA standard\footnote{http://www.fao.org/gsfaonline/index.html?lang=en}, the food from our dataset and existing typical ones mainly belongs to the following 11  categories: Meat, Cereals, Vegetables, Fish, Fruits, Dairy, Bakery, Fats, Confectionary, Beverages and Eggs. We can see that for most of food types, the number of food categories from  ISIA Food-500 is larger than these existing datasets. Furthermore, some food types are covered in ISIA Food-500, but missing in other ones, such as Dairy and Beverages.

\section{Framework}
\begin{figure*}
	\centering
	\includegraphics[width=0.750\textwidth]{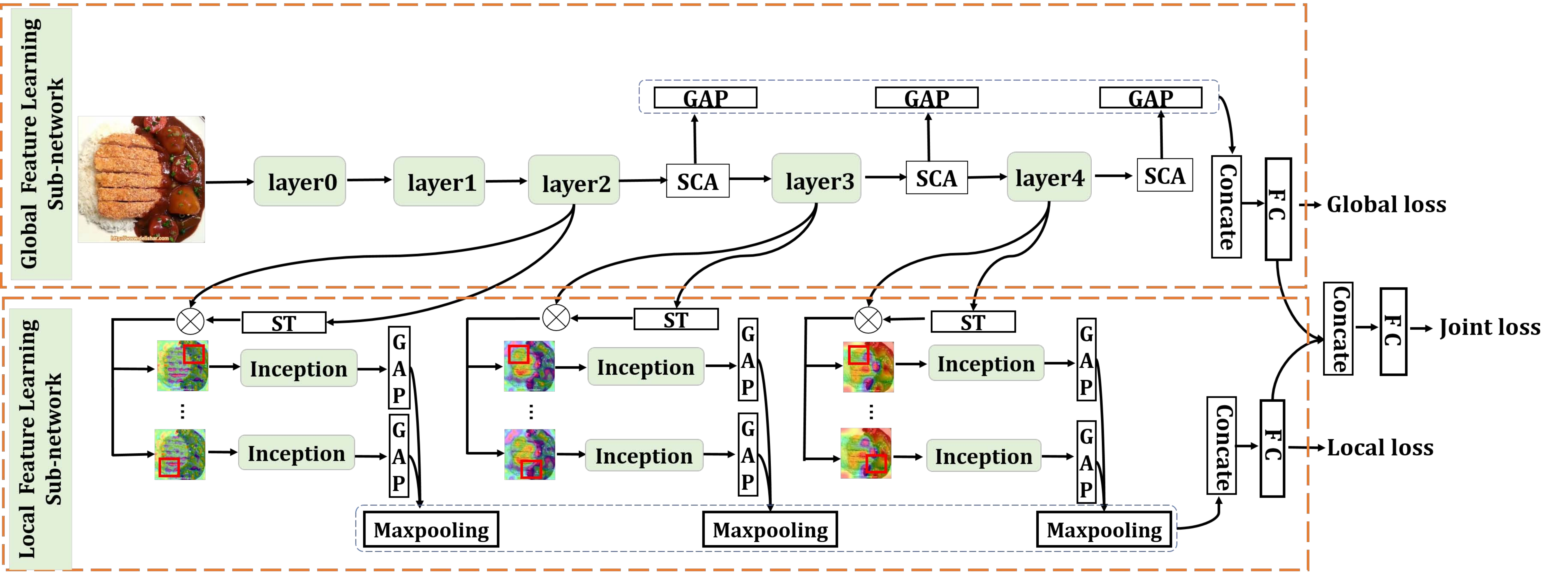}
	\caption{The proposed framework. GAP: Global Average Pooling layer. SCA: Spatial-Channel Attention. ST: Spatial Transformer. FC: Full-Connected layer.}
	\label{framework}
\end{figure*}
Fig.~\ref{framework} illustrates the proposed  Stacked Global-Local Attention Network (SGLANet), which can jointly learn complementary   global and local  features for food recognition. SGLANet mainly consists of  two components, namely \textbf{Glo}bal \textbf{F}eature \textbf{L}earning \textbf{S}ub-network (GloFLS) and \textbf{Loc}al-\textbf{F}eature \textbf{L}earning \textbf{S}ub-network (LocFLS). GloF-LS first   adopts  hybrid Spatial-Channel Attention (SCA) to obtain more discriminative  features from each layer of the network, and then aggregates a set of  features from these  layers to capture different types of global level features, such as  shape and texture cues about food. LocFLS  adopts cascaded STs to localize different local  regions  for each layer, and then  aggregates fused features with different  regions from different layers into final local feature representation. Finally, SGLANet fuses both global and local features for food recognition. In addition, SGLANet is trained  with different types of losses, including  global loss, local loss and joint loss in an end-to-end fashion to  maximize their complementary benefit in terms of the discriminative power.
{\Large {\LARGE }}\subsection{GloFLS}
Given the whole input image, GloFLS first learns more discriminative features via hybrid Spatial-Channel Attention (SCA) for each layer, and then  aggregates these discriminative features from different layers into global level representations via multi-layer feature fusing. Considering  features extracted from different layers contain low-level, mid and high ones, GloFLS  can capture various types of global level features, such as  shape, texture and edge cues  about food.

\textbf{Spatial-Channel Attention~(SCA)}
The combination of both spatial and channel attention can  capture discriminative features comprehensively from different dimensions, and thus have been successfully applied in many tasks, such as  image captioning~\cite{Chen-SCA-CNN-CVPR2017} and person ReID~\cite{Li-HAN-CVPR2018}. Different from these works, we apply SCA to the food recognition task by capturing food-oriented discriminative features.

The input to a SCA module is a 3-D tensor \textbf{X}$^{l}${$\in$}{R} $^{h\times w \times c}$ with width $w$, height $h$, channels $c$ and  the layer of GloFLS ${l}$, respectively. The output of this module is  a saliency weight map \textbf{A}$^{l}${$\in$}{R} $^{h\times w \times c}$ of the same size as \textbf{X}. We calculate \textbf{A}$^{l}${$\in$}{R} for SCA learning~\cite{Li-HAN-CVPR2018}:
\begin{equation}
\begin{aligned}
&\textbf{A}^{l}=\textbf{S}^{l} \times \textbf{C}^{1}
\end{aligned}
\end{equation}
where  \textbf{S}$^{l}${$\in$}{R}$^{h\times w \times 1}$ and \textbf{C}$^{l}${$\in$}{R}$^{1\times 1 \times c}$ mean spatial and channel attention maps, respectively.

The Global Averaging Pooling (GAP) is used to calculate the spatial attention as follows:
\begin{equation}
\begin{aligned}
&\textbf{S}^{l}=\frac{1}{c}\sum\limits^{c}_{i=1}\textbf{X}^{l}_{1:h,1:w:i}
\end{aligned}
\end{equation}
The channel attention from the squeeze-and-excitation block~\cite{Jie2017Squeeze} is computed as follows:
\begin{equation}
\begin{aligned}
&\mathbf{C}_{1}^{l}=\frac{1}{h \times w}\sum\limits^{h}_{i=1}\sum\limits^{w}_{j=1}\mathbf{X}^{l}_{i,j,1:c}\\
&\mathbf{C}^{l}={ReLU}(\mathbf{M}^{ca}_{2} \times {\rm Relu}(\mathbf{M}^{ca}_{1}\mathbf{C}_{1}^{l}))
\end{aligned}
\end{equation}
where \textbf{M}$^{ca}_{1}$ $\in$ \textbf{R}$^{\frac{c}{r} \times c}$
and \textbf{M}$^{ca}_{2}$ $\in$ \textbf{R}$^{c \times \frac{c}{r}}$
represent the parameter matrix of 2 conv layers respectively, and $r$ denotes the bottleneck reduction rate.

\textbf{Multi-Layer Feature Fusing} By extracting attentional features from multiple layers, we can obtain low, mid and high-level features, which include various types of global features, such as texture, shape and edge information ~\cite{Yang-MSR-ICCV2015}. Such global features are  important cues for food recognition. Therefore, we aggregate discriminative  attentional features from different layers  into global level feature representation for food recognition via  a concatenation layer and a fully connected layer.

\subsection{LocFLS}
LocFLS localizes discriminative regions with different positions and scales  to capture local features. It uses stacked STs~\cite{Jaderberg-STN-NIPS2015} to localize regions from different layers. For each layer, one inception block is introduced to extract regional features, and followed by a global average pooling layer and a max-pooling layer for fusing these regional features. The features from each layer are  fused to final local features via a concatenation layer and a fully connected layer.

\textbf{Spatial Transformer (ST)}
For each layer, we adopt   ST   to locate latent  $T$ regions, and model this regional attention by a transformation matrix as:
\begin{equation}
\begin{aligned}
&\textbf{A}^{l}=
\left[
\begin{matrix}
s_{h} & 0 & t_{x} \\
0 & s_{w} & t_{y} \\
\end{matrix}
\right]
\end{aligned}
\end{equation}
which allows for image cropping, translation, and isotropic scaling operations by varying two scale factors ($s_{h}$ , $s_{w}$) and  2-D spatial position ($t_{x}$, $t_{y}$).

\subsection{Learning with Multiple Losses    }

SGLANet is jointly optimized  by three types of losses, i.e., joint loss $L_{Joi}$, global  loss $L_{Glo}$, and local loss $L_{Loc}$ respectively, leading to  the final loss function:
\begin{equation}\label{MTL}
\begin{aligned}
L =  L_{Joi} + \gamma_{1}L_{Glo} + \gamma_{2}L_{Loc}
\end{aligned}
\end{equation}
where $\gamma_{1}$ and $\gamma_{2}$ are balance parameters, and  the cross-entropy classification loss function is used  for all three types of losses.

Such  learning with different types of losses can  maximize their complementary benefit in terms of the discriminative power.

\section{Experiment}
\subsection{Experimental Setup}
Our model is implemented on the Pytorch platform. The images are resized to 224$\times$224. The models are optimized using stochastic gradient descent with a batch size of 80 and momentum of 0.9. The learning rate is set to $10^{-2}$ initially and divided by 10 after 30 epochs. For GloFLS, we selected SENet~\cite{Jie2017Squeeze} as the backbone, and the  bottleneck reduction rate $r=16$.  For  LocFLS, we selected simple Inception-B unit  as  basic building block. For each layer, $T=4$ and the scale of ST is fixed as $s_{h}=s_{w}=0.5$. We set $\gamma_{1}=\gamma_{2}=0.5$  in Eq.~\ref{MTL}.  Top-1 accuracy (Top-1 acc.) and Top-5 accuracy (Top-5 acc.)  are used as  evaluation metrics.
\subsection{Experiment on ISIA Food-500}
ISIA Food-500 is divided into 60\%, 10\% and 30\% images for training, validation and testing, respectively. All the experiments adopt a single centered crop (1-crop) at test time in the defaulting setting.

\textbf{Ablation Study}
We first evaluated the effect of each individual component in GloFLS in Table~\ref{ablation_GloFLS_food500}. It shows that: (1) Any of two components in isolation brings recognition performance gain; (2) The combination of SCA and Multi-scale gives further accuracy boost, which suggests the complementary effect. We then evaluated the effect of joint  global and local feature learning by comparing their individual global and local features. Table ~\ref{ablation_food500} shows that a  performance gain is obtained in Top-1 accuracy by joining two representations, which validates the complementary effect of jointly learning global and local features from GloFLS and LocFLS.
\begin{table}[!t]
	\caption{Evaluating individual modules in GloFLS on ISIA Food-500 (\%).}
	\label{ablation_GloFLS_food500}
	\begin{center}
		\begin{tabular}{|c|c|c|}
			\hline
			\textbf{Method}& Top-1 acc. & Top-5 acc.\\
			\hline		
			SENet-154& 63.83 & 88.61 \\
			\hline
			SENet-154+SCA& 64.42&89.05\\
			\hline
			SENet-154+Multi-scale&64.60&89.08\\
			\hline
			GloFLS&\textbf{64.63}&\textbf{89.14}\\
			\hline
		\end{tabular}
	\end{center}
\end{table}

\begin{table}[!t]
	\caption{Ablation experiments on ISIA Food-500 with global \& local-level features (\%).}
	\label{ablation_food500}
	\begin{center}
		\begin{tabular}{|c|c|c|}
			\hline
			\textbf{Method}& Top-1 acc. & Top-5 acc.\\
			\hline
			%			\multirow{2}*{\textbf{Method}} &
			%			\multicolumn{2}{|c|}{Public Test} &\multicolumn{2}{|c|}{Private Test}\\
			%			\cline{2-5}& Top-1 acc.& Top-5 acc.& Top-1 acc.& Top-5 acc.\\
			%			\hline			
			GloFLS&64.63&\textbf{89.14}\\
			\hline
			LocFLS& 64.10&88.86\\
			\hline
			SGLANet&\textbf{64.74}&89.12\\
			\hline
		\end{tabular}
	\end{center}
\end{table}

We finally evaluate the effect of different losses as shown in Table~\ref{ablation_MultiLoss_food500}. The experimental results demonstrate that we obtain the best recognition performance when different losses are utilized. The reason is that different loss functions can regulate the deep network from different aspects and work together to improve the recognition performance. Another observation is that the performance with one additional  loss does not improve the performance compared with the baseline without both global and local losses. The probable reason is that the performance improvement needs joint work from two losses.

\begin{table}[!t]
	\caption{Evaluating individual losses on ISIA Food-500 (\%).}
	\label{ablation_MultiLoss_food500}
	\begin{center}
		\begin{tabular}{|c|c|c|}
			\hline
			\textbf{Method}& Top-1 acc. & Top-5 acc.\\
			\hline		
			$\lambda_{1}=\lambda_{2}=0$&64.16  &88.94  \\
			\hline
			$\lambda_{1}=0$, $\lambda_{2}=0.5$&63.95 &88.57\\
			\hline
			$\lambda_{1}=0.5$, $\lambda_{2}=0$&64.02 &88.59\\
			\hline
			$\lambda_{1}=0.5$, $\lambda_{2}=0.5$&\textbf{64.74}&\textbf{89.12} \\
			\hline
		\end{tabular}
	\end{center}
\end{table}

\textbf{Comparisons with State-of-the-Art}
We evaluated SGLANet against different baseline  methods on Table \ref{Performance_com_food500}. These baselines include not only various typical deep networks, such as VGG16 and SENet, but also some recently proposed fine-grained methods, such as NTS-NET~\cite{Yang-L2Nav-ECCV2018} and WS-DAN~\cite{Hu_2019_CVPR}. Note that for these fine-grained methods, we followed the same setting in their mentioned papers. We observed that the  performance superiority of SGLANet over all the state-of-the-arts in both Top-1 accuracy and Top-5 accuracy. Compared with best baseline  SENet-154, there is the performance improvement of about 0.9 percent  in Top-1 accuracy for the test set. These results validate the  advantage of joint global and local feature learning of SGLANet.

\begin{table}[!t]
	\caption{Performance comparison  on ISIA Food-500 (\%).}
	\label{Performance_com_food500}
	\begin{center}
		\begin{tabular}{|c|c|c|}
			\hline
			\textbf{Method}& Top-1 acc. & Top-5 acc.\\
			%				\multirow{2}*{\textbf{Method}} &
			%				\multicolumn{2}{|c|}{Public Test} &\multicolumn{2}{|c|}{Private Test}\\
			%				\cline{2-5}& Top-1 Acc.& Top-5 acc.& Top-1 acc.& Top-5 acc.\\
			%				\hline	
			\hline		
			VGG-16~\cite{Szegedy-GDC-CVPR2015}& 55.22 & 82.77 \\
			\hline
			GoogLeNet~\cite{Meyers-Im2Calories-ICCV2015}& 56.03 &83.42\\
			\hline
			ResNet-152~\cite{He-DRL-CVPR2016}& 57.03& 83.80  \\
			\hline		
			WRN-50~\cite{Zagoruyko-WRN-BMVC2016}& 60.08& 85.98\\
			\hline
			DenseNet-161~\cite{huang2017densely}&  60.05& 86.09\\
			\hline
			NAS-NET~\cite{Zoph-NAS-NET-CVPR2018}& 60.66 & 86.38\\
			\hline
			SE-ResNeXt101\_32x4d~\cite{Jie2017Squeeze}& 61.95 & 87.54\\
			\hline
			
			NTS-NET~\cite{Yang-L2Nav-ECCV2018}& 63.66 &  88.48\\
			\hline
			WS-DAN~\cite{Hu_2019_CVPR}& 60.67 & 86.48 \\
			\hline
			DCL~\cite{Chen_2019_CVPR}& 64.10 & 88.77 \\
			\hline
			SENet-154~\cite{Jie2017Squeeze}& 63.83 & 88.61 \\
			\hline
			SGLANet&\textbf{64.74}&\textbf{89.12}\\
			\hline
		\end{tabular}
	\end{center}
\end{table}

\begin{figure}
	\centering
	\includegraphics[width=0.45\textwidth]{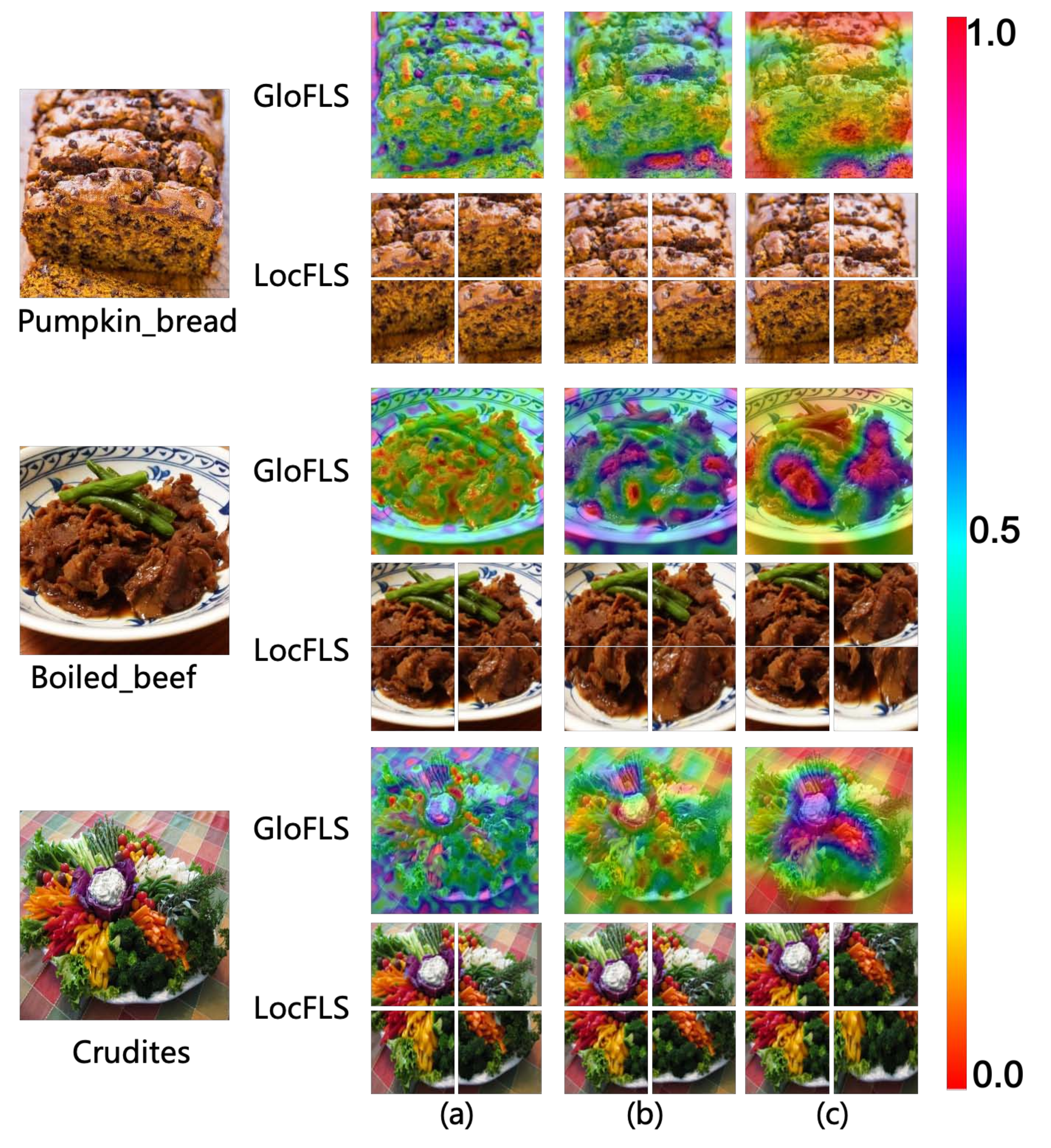}
	\caption{Visualization of SCA in GloFLS and STs in LocFLS from (a) The $2^{\rm th}$ layer,(b) The $3^{\rm th}$ layer and (c) The $4^{\rm th}$ layer.}
	\label{Visualization}
\end{figure}
\begin{figure}[!t]
	\centering
	\subfigure[The 10 best performing classes]{\includegraphics[width=0.5\textwidth]{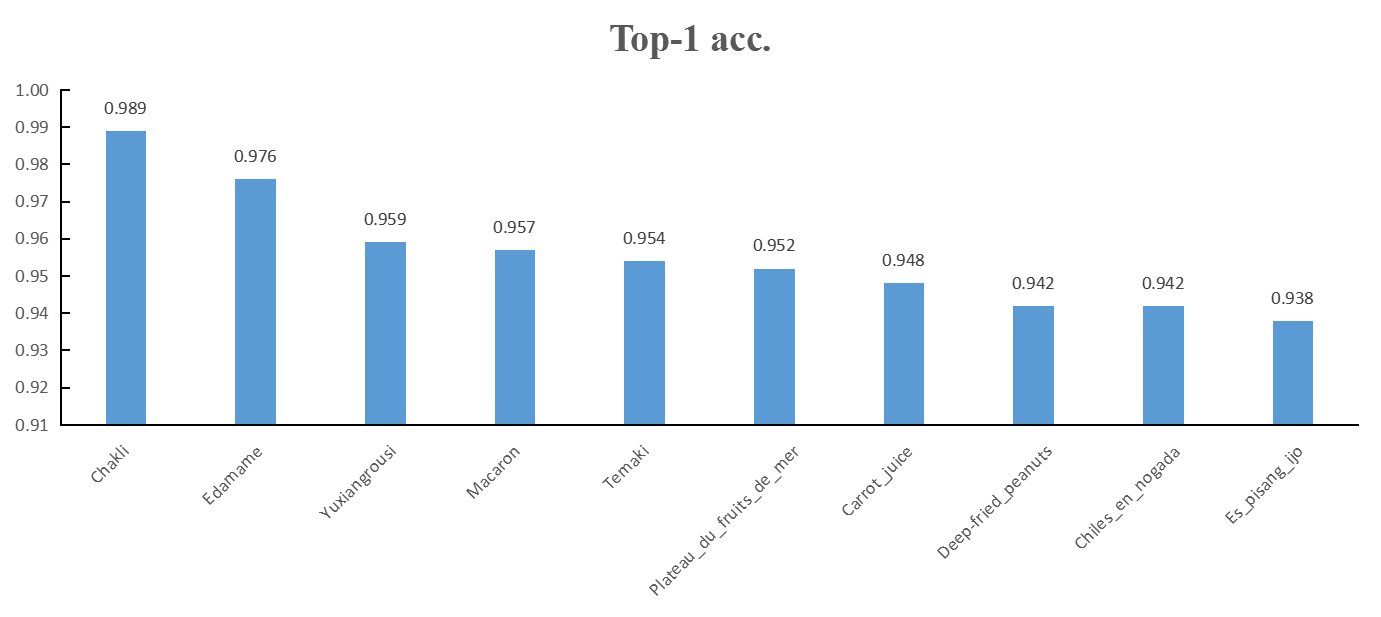}
		\label{fig_first_case}}
	\vfill
	\subfigure[The 10 worst performing classes]{\includegraphics[width=0.5\textwidth]{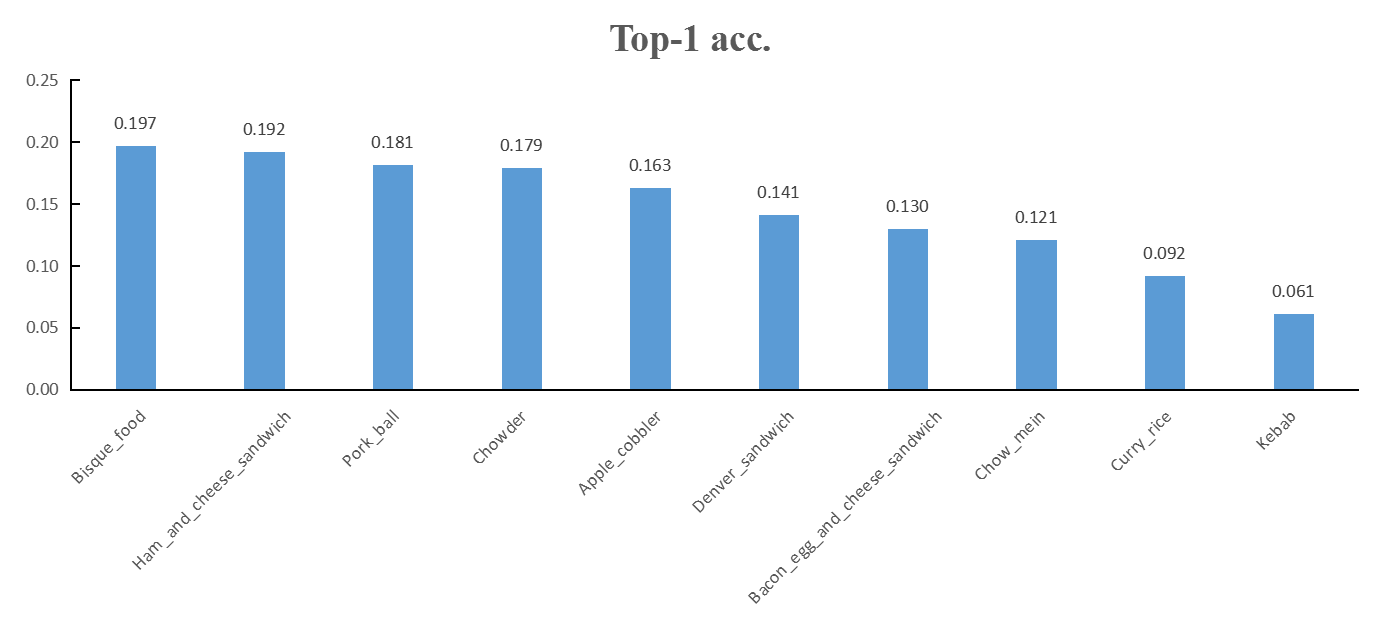}
		\label{fig_second_case}}
	\caption{Selected categories from (a)The 10 best and (b)The 10 worst performing classes.}
	\label{Top_10_Acc}
\end{figure}

\begin{figure}
	\centering
	\includegraphics[width=0.40\textwidth]{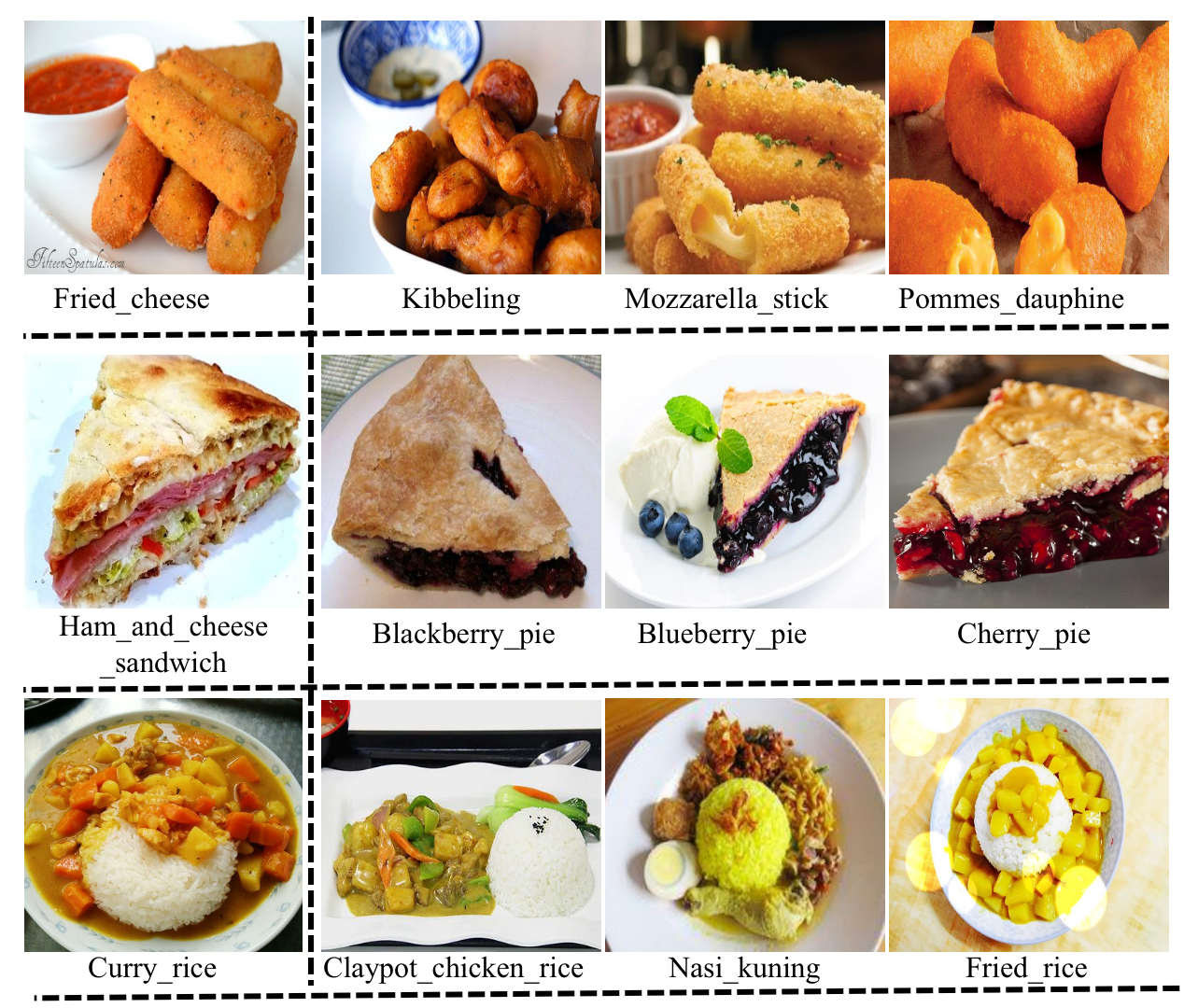}
	\caption{Some confused classes, where the first column denotes some classes from the 10 worst performing classes and for each class, 3 more confused classes  are listed for each row.}
	\label{wrong_result_analysis}
\end{figure}

\textbf{Visualization of GloFLS and LocFLS}
We visualized both SCA  from GloFLS and STs from LocFLS at three different
layers of SGLA-Net. Fig. \ref{Visualization} shows: (1) in GloFLS, SCA captures different global level features at different layers, such as shape information for Boiled$\_$beef and texture information from Pumpkin$\_$bread. Meanwhile, with increased depth of SGLANet, SCA captures more focused and discriminative features (2) in LocFLS, STs  capture different local regions with less background at different layers from LocFLS, such as Crudites. This again verified complementary effect of joint global and local feature learning.

\textbf{Qualitative Analysis} We selected 20 classes in the test phase to further evaluate our method. Particularly, we listed the Top-1 accuracy of both 10 best and 10 worst performing classes in Fig. \ref{Top_10_Acc}. We can observe that some categories can be easily recognized, such as Chakli and Edamame, and their Top-1 accuracy is above 97\%. However, there are some categories, which are very hard to recognize, such as Curry$\_$rice and kebab, and their Top-1 accuracy is below 10\%. We further demonstrate some challenging recognized examples from the 10 worst  performing classes, and Fig. \ref{wrong_result_analysis} shows  that too small inter-class variations is the main reason for bad performance. We have shown that existing methods are far from tackling large-scale recognition task with high accuracy like ImageNet, pointing to exciting future directions.

\begin{table}[htbp]
	\caption{Performance comparison  on ETHZ Food-101 (\%).}
	\label{recognition_performance_food101}
	\begin{center}
		\begin{tabular}{|c|c|c|c|}
			\hline
			\textbf{Method}&Setting& Top-1 acc. & Top-5 acc.\\
			\hline
			AlexNet-CNN~\cite{Bossard-Food101-ECCV2014}&1-crop&56.40&-\\
			SELC~\cite{Martinel2016A} &1-crop&55.89&-\\
			ResNet-152+SVM-RBF~\cite{mcallister2018combining}&1-crop&64.98&-\\
			DCNN-FOOD~\cite{Yanai-FIRDCNN-ICME2015}&1-crop&70.41&-\\
			LMBM~\cite{Wu-LMBM-MM2016}&1-crop&72.11&- \\
			Ensemble Net~\cite{Pandey2017FoodNet}&1-crop&72.12&91.61\\
			GoogLeNet~\cite{Ao2015Adapting}&1-crop&78.11&- \\
			DeepFOOD~\cite{Liu2016DeepFood}&1-crop&77.40&93.70\\
			ILSVRC~\cite{Bola2017Simultaneous}&1-crop&79.20&94.11\\
			WARN~\cite{Rodriguez-PATA-TMM2019}&1-crop&85.50&- \\
			CNNs Fusion(I$_{2}$)~\cite{Aguilar2017Food}&1-crop&86.71&-\\
			Inception V3~\cite{Hassanne-FIRDCN-MM2016}&1-crop&88.28&96.88\\
			SENet-154~\cite{Jie2017Squeeze}&1-crop& 88.62 &97.57 \\
			WRN~\cite{Martinel-WSR-WACV2018}&10-crop&88.72&97.92\\
			SOTA\cite{Kornblith2018Do}&1-crop&90.00&- \\
			DLA\cite{Yu-DLA-CVPR2018}&1-crop&90.00&- \\
			WISeR~\cite{Martinel-WSR-WACV2018} &10-crop&90.27&\textbf{98.71} \\
			IG-CMAN~\cite{Min-IGCMAN-ACMMM2019}&1-crop &90.37&98.42 \\
			PAR-Net~\cite{Qiu-MDFR-BMVC2019}&1-crop&89.30&- \\
			PAR-Net~\cite{Qiu-MDFR-BMVC2019}&10-crop&90.40&- \\
			Inception-Resnet-v2 SE~\cite{Cui-LSFGC-CVPR2018}&1-crop&90.40&-\\
			MSMVFA~\cite{Min-MSMVFA-TIP2019} &1-crop&{90.59}&{98.25 }\\
			\hline
			SGLANet &1-crop&89.69&98.01\\
			SGLANet &10-crop&90.33&98.20\\
			SGLANet(Pretrained)&1-crop&90.47 &98.21\\
			SGLANet(Pretrained)&10-crop&\textbf{90.92}&98.24\\
			\hline
		\end{tabular}
	\end{center}
\end{table}

\subsection{Experiment on Other Benchmarks}
We further conduct extensive  evaluation on other two popular food benchmark datasets to verify the effectiveness of  our approach, and also assessed the generalizability of our model learned using ISIA Food-500 to the two datasets. Considering some evaluations from existing works are conducted in the setting of 10-crop test, we  show the experimental results of our method in the setting of  both   1-crop and  10-crop at test time.
\begin{table} [t]
	\caption{Performance comparison on Vireo Food-172 (\%).}
	\label{food-172 dataset_Results}% Give a unique label
	\centering
	\begin{tabular}{|c|c|c|c|}
		\hline
		\textbf{Method}&Setting&Top-1 acc.&Top-5 acc.\\
		\hline
		AlexNet&1-crop&64.91 &85.32 \\
		VGG-16~\cite{Szegedy-GDC-CVPR2015}&1-crop&80.41 &94.59 \\
		DenseNet-161~\cite{huang2017densely}&1-crop&86.93 &97.17 \\
		MTDCNN(VGG-16)~\cite{Chen-DIRCRR-MM2016}&1-crop&82.06 &95.88 \\
		MTDCNN(DenseNet-16)~\cite{Chen-DIRCRR-MM2016}&1-crop&87.21 &97.29 \\
		SENet-154~\cite{Jie2017Squeeze}&1-crop& 88.71 & 97.74\\
		PAR-Net~\cite{Qiu-MDFR-BMVC2019}&1-crop&89.60&- \\
		PAR-Net~\cite{Qiu-MDFR-BMVC2019}&10-crop&90.20&- \\
		IG-CMAN~\cite{Min-IGCMAN-ACMMM2019}&1-crop&90.63 &\textbf{98.40} \\
		MSMVFA~\cite{Min-MSMVFA-TIP2019}&1-crop&90.61 &98.31 \\
		\hline
		SGLANet&1-crop&89.88 &97.83\\
		SGLANet&10-crop& 90.30&98.03\\
		SGLANet(Pretrained)&1-crop&90.78&98.16\\
		SGLANet(Pretrained)&10-crop&\textbf{90.98} &98.35\\
		\hline
	\end{tabular}
\end{table}	

\textbf{Experiments on ETHZ Food-101}
ETHZ Food-101 contains 101,000 images from 101 food categories. There are 1,000 images including 750 training images and 250 test images for each category~\cite{Bossard-Food101-ECCV2014}.  We evaluated SGLANet against existing methods on Food-101. Table \ref{recognition_performance_food101} shows that our method exceeds many baseline methods except some ones, such as MSMVFA~\cite{Min-MSMVFA-TIP2019}, IG-CMAN~\cite{Min-IGCMAN-ACMMM2019} and Inception-Resnet-v2 SE~\cite{Cui-LSFGC-CVPR2018} under the 1-crop test setting. The reason is that  MSMVFA and IG-CMAN  require multiple stages training for feature extraction and  introduced additional ingredient information as the supervision. Inception-Resnet-v2 SE  used  additional data and adopted transfer learning method. When we used the pretrained model on ISIA Food-500, namely SGLANet(Pretra-ined), there is the performance improvement of about 0.8 percent and 0.6 percent in Top-1 accuracy on 1-crop and 10-crop test respectively. These results also verify the generalization of models learned using ISIA Food-500.

\textbf{Experiments on Vireo Food-172}
Vireo Food-172 consists of 110,241 food images from 172 categories. In each food category, 60\%, 10\%, 30\% of images are randomly selected for training, validation and testing, respectively~\cite{Chen-DIRCRR-MM2016}. Table \ref{food-172 dataset_Results} shows experimental results on Vireo Food-172. We can see that the performance from SGLANet is better than many baselines, except that few ones, such as IG-CMAN. This is because that these methods, such as IG-CMAN  introduced additional ingredient information for food recognition. In addition, these methods generally need multi-stage feature learning.  When we fine-tuned  SGLANet pre-trained on ISIA Food-500, there is the performance improvement of about 0.9 percent and 0.7 percent in Top-1 accuracy on 1-crop and 10-crop test respectively, and achieved the best performance under the 1-crop setting.  These results again demonstrate the generalization of our model learned using ISIA Food-500.

\section{Conclusions}
In this paper, we present a new large-scale dataset ISIA Food-500 with  larger data volume, larger category coverage, and higher diversity compared with existing typical datasets. We then propose a stacked  global-local attention network to jointly exploit complementary global and local features via the designed two subnetworks for food recognition. Extensive evaluation on ISIA Food-500 and another two benchmark datasets have verified its effectiveness, and thus can be considered as one strong baseline.

Future work includes: (1) We  are  expanding ISIA Food-500 data-set, and aim to  complete the construction of about 1.5 million food images spread over about 2,000 food categories. We expect it will serve as a new challenge to  train high-capacity models for large-scale food recognition in the multimedia community. (2) We plan to collect rich attribute information, e.g., ingredients, cooking instructions and flavor information~\cite{Weiqing-DRAF-MM2017} to support  multimodal  food recognition.

\begin{acks}
This work was supported by the National Natural Science Foundation of China under Grant 61972378, 61532018, U1936203, U19B2040. This research was also supported by Meituan-Dianping Group.
\end{acks}

%
% The next two lines define the bibliography style to be used, and the bibliography file.
\bibliographystyle{ACM-Reference-Format}
\balance
\bibliography{sigproc}

\end{document}